# Mixtures of Deterministic-Probabilistic Networks and their AND/OR Search Space

**Rina Dechter** and **Robert Mateescu**
School of Information and Computer Science
University of California, Irvine, CA 92697-3425
{dechter, mateescu}@ics.uci.edu

## Abstract

The paper introduces *mixed networks*, a new framework for expressing and reasoning with probabilistic and deterministic information. The framework combines belief networks with constraint networks, defining the semantics and graphical representation. We also introduce the AND/OR search space for graphical models, and develop a new linear space search algorithm. This provides the basis for understanding the benefits of processing the constraint information separately, resulting in the pruning of the search space. When the constraint part is tractable or has a small number of solutions, using the mixed representation can be exponentially more effective than using pure belief networks which model constraints as conditional probability tables.

## 1 INTRODUCTION

Modeling real-life decision problems requires the specification and reasoning with probabilistic and deterministic information. The primary approach developed in artificial intelligence for representing and reasoning with partial information under conditions of uncertainty is Bayesian networks. They allow expressing information such as "if a person has flu, he is likely to have fever." Constraint networks and propositional theories are the most basic frameworks for representing and reasoning about deterministic information. Constraints often express resource conflicts frequently appearing in scheduling and planning applications, precedence relationships (e.g., "job 1 must follow job 2") and definitional information (e.g., "a block is clear iff there is no other block on top of it"). Most often the feasibility of an action is expressed using a deterministic rule between the pre-conditions (constraints) and post-conditions that must hold before and after executing an action (e.g., STRIPS for classical planning).

The two communities of probabilistic networks and constraint networks matured in parallel with only minor interaction. Nevertheless some of the algorithms and reasoning principles that emerged within both frameworks, especially those that are graph-based, are quite related. Both frameworks can be viewed as graphical models, a popular paradigm for knowledge representation.

Researchers within the logic-based and constraint communities have recognized for some time the need for augmenting deterministic languages with uncertainty information, leading to a variety of concepts and approaches such as non-monotonic reasoning, probabilistic constraint networks and fuzzy constraint networks. The belief networks community started only recently to look into mixed representation [Poole1993, Ngo & Haddawy1977, Dechter & Larkin2001] perhaps because it is possible, in principle, to capture constraint information within belief networks [Pearl1988].

Indeed, constraints can be embedded within belief networks by modeling each constraint as a Conditional Probability Table (CPT). One approach is to add a new variable for each constraint that is perceived as its *effect* (child node) in the corresponding causal relationship and then to clamp its value to *true* [Pearl1988]. While this approach is semantically coherent and complies with the acyclic graph restriction of belief networks, it adds a substantial number of new variables, thus cluttering the problem's structure. An alternative approach is to designate one of the arguments of the constraint as a child node (namely, as its effect). This approach, although natural for functions (the arguments are the causes or parents and the function variable is the child node), is quite contrived for general relations (e.g., $x + 6 \neq y$). Such constraints may lead to cycles, which are disallowed in belief networks. Furthermore, if a variable is a child node of two different CPTs (one may be deterministic and one probabilistic) the belief network definition requires that they be combined into one CPT.

The main shortcoming, however, of any of the above integrations is computational. Constraints have special properties that render them attractive computationally. When con-straints are disguised as probabilistic relationships, their computational benefits are hard to exploit. In particular, the power of constraint inference and constraint propagation may not be brought to bear.

Therefore, we propose a framework that combines deterministic and probabilistic networks, called *mixed network*. Specifically, we propose a mixed network framework in which the identity of the respective relationships, as constraints or probabilities, will be maintained explicitly, so that their respective computational power and semantic differences can be vivid and easy to exploit. The mixed network approach allows two distinct representations: causal relationships that are directional and normally (but not necessarily) quantified by CPTs and symmetrical deterministic constraints. The proposed scheme's value is in providing: 1) semantic coherence; 2) user-interface convenience (the user can relate better to these two pieces of information if they are distinct); and most importantly, 3) computational efficiency.

## 2  PRELIMINARIES AND BACKGROUND

**Reasoning graphical model**  A reasoning graphical model is a triplet $\mathcal{R} = (X, D, F)$ where $X$ is a set of variables, $X = \{X_1, \ldots, X_n\}$, $D = \{D_1, \ldots, D_n\}$ is the set of their respective finite domains and $F = \{F_1, \ldots, F_t\}$ is a set of real-valued functions, defined over subsets of $X$. The *primal graph* of a reasoning problem has a node for each variable, and any two variables appearing in the same function's scope are connected. The *scope* of a function is its set of arguments.

**Belief networks**  A belief network can be viewed as an instance of a reasoning graphical model. In this case the set of functions $F$ is denoted by $P = \{P_1, \ldots, P_n\}$ and represents a set of conditional probability tables (CPTs): $P_i = P(X_i|pa_i)$. $pa_i$ are the parents of $X_i$. When the CPTs entries are "0" or "1" only, they are called *deterministic or functional CPTs*. The associated directed graph $G$, drawn by pointing arrows from parents to children, should be acyclic. We also denote belief networks by $\mathcal{B} = (X, D, G, P)$. The belief network represents a probability distribution over $X$ having the product form $P_\mathcal{B}(\bar{x}) = P(x_1, \ldots, x_n) = \Pi_{i=1}^n P(x_i | x_{pa_i})$ where an assignment $\bar{x} = (X_1 = x_1, \ldots, X_n = x_n)$ is abbreviated to $\bar{x} = (x_1, \ldots, x_n)$ and where $x_S$ or $x[S]$ denote the restriction of a tuple $x$ over a subset of variables $S$. An evidence set $e$ is an instantiated subset of variables. We use upper case letters for variables and nodes in a graph and lower case letters for values in a variable's domain. The *moral graph* of a directed graph is the undirected graph obtained by connecting the parent nodes of each variable and eliminating direction. Given a directed graph $G$, *the ancestral graph relative to a subset of nodes* $X$ is the undirected graph obtained by taking the subgraph of $G$ that contains $X$ and all their non-descendants, and moralizing the graph.

**Constraint networks**  A constraint network can also be viewed as an instance o a reasoning graphical model. In this case the functions are denoted by $C = \{C_1, \ldots, C_t\}$, and the constraint network is denoted by $\mathcal{R} = (X, D, C)$. Each constraint is a pair $C_i = (S_i, R_i)$, where $S_i \subseteq X$ is the scope of the relation $R_i$ defined over $S_i$, denoting the allowed combinations of values. The associated graph $G$ of a constraint network $\mathcal{R}$ is its primal graph. We say that $\mathcal{R}$ represents its set of solutions, $\rho$, or $\rho(\mathcal{R})$. A particular example of constraint networks is CNF, in which the variables are boolean (binary domains) and the constraints are boolean formulas. In this case the network is given as formula in conjunctive normal form.

**Induced-graphs and induced width**  An *ordered graph* is a pair $(G, d)$ where $G$ is an undirected graph, and $d = X_1, \ldots, X_n$ is an ordering of the nodes. The *width of a node* in an ordered graph is the number of the node's neighbors that precede it in the ordering. The *width of an ordering* $d$, denoted $w(d)$, is the maximum width over all nodes. The *induced width of an ordered graph*, $w^*(d)$, is the width of the induced ordered graph obtained as follows: nodes are processed from last to first; when node $X$ is processed, all its preceding neighbors are connected. The *induced width of a graph*, $w^*$, is the minimal induced width over all its orderings. The *tree-width* of a graph is the minimal induced width.

**Tasks**  The primary queries over belief networks are: *belief updating*, evaluating the posterior probability of each singleton proposition given some evidence; *most probable explanation* (MPE), finding a complete assignment to all variables having maximum probability given the evidence and *maximum a posteriori hypothesis* (MAP), which calls for finding the most likely assignment to a subset of hypothesis variables given the evidence. The primary queries over constraint networks are to decide if the network is consistent and if so, to find one, some or all solutions.

## 3  MIXING PROBABILITIES WITH CONSTRAINTS

DEFINITION **1 (mixed networks)**  *Given a belief network $\mathcal{B} = (X, D, G, P)$ that expresses the joint probability $P_\mathcal{B}$ and given a constraint network $\mathcal{R} = (X, D, C)$ that expresses a set of solutions $\rho$, a mixed network based on $\mathcal{B}$ and $\mathcal{R}$ denoted $\mathcal{M}_{(\mathcal{B},\mathcal{R})} = (X, D, G, P, C)$ is created from the respective components of the constraint network and the belief network as follows. The variables $X$ and their domains are shared, (we could allow non-common variables and take the union), and the relationships include the CPTs in $P$ and the constraints in $C$. The mixed network may be inconsistent, or if it is consistent it expresses the conditional probability $P_\mathcal{M}(X)$:*

$$P_\mathcal{M}(\bar{x}) = \begin{cases} P_\mathcal{B}(\bar{x} \mid \bar{x} \in \rho), & if \ \bar{x} \in \rho \\ 0, & otherwise. \end{cases}$$

Belief updating, MPE and MAP queries can be extended to mixed networks straightforwardly. They are well defined relative to the mixed probability $P_\mathcal{M}$, when the constraint portion is consistent. An additional relevant query over a mixed network is to find the probability that a random tuple satisfies the constraint query, namely $P_\mathcal{B}(\bar{x} \in \rho)$.

**The auxiliary network** We now define the belief network that expresses constraints as pure CPTs.

DEFINITION **2 (auxiliary network)** *Given a mixed network $M_{(\mathcal{B},\mathcal{R})}$ we define the auxiliary network $S_{(\mathcal{B},\mathcal{R})}$ to be a belief network that has new auxiliary variables as follows. For every constraint $C_i = (S_i, R_i)$ in $\mathcal{R}$, we add the auxiliary variable $A_i$ that has a domain of two values, $\{0, 1\}$. There is a CPT defined over $A_i$ whose parent variables are $S_i$, defined as follows:*

$$P(A_i=1 \mid t_{S_i}) = \begin{cases} 1, & if \ t \in R_i \\ 0, & otherwise. \end{cases}$$

$S_{(\mathcal{B},\mathcal{R})}$ is a belief network that expresses a probability distribution $P_S$. It is easy to see that,

**Proposition 1** *Given a mixed network $M_{(\mathcal{B},\mathcal{R})}$ and an associated auxiliary network $S = S_{(\mathcal{B},\mathcal{R})}$, then: $P_\mathcal{M}(\bar{x}) = P_S(\bar{x}|A_1=1, ..., A_t=1)$.*

One source of determinism in the context of belief networks may arise because we have deterministic queries or complex evidence description. Both reduce to *CNF or Constraint Probability Evaluation (CPE)*.

DEFINITION **3 (CPE)** *Given a mixed network $M_{(\mathcal{B},\mathcal{R})}$, where the belief network $(X, D, G, P)$ is defined over variables $X = \{X_1, ..., X_n\}$ and where the constraint portion is a either a set of relational constraints or a CNF query ($\mathcal{R} = \varphi$) over a subset $Q = \{Q_1, ...Q_r\}$, where $Q \subseteq X$, the* Constraint, *respectively* CNF, Probability Evaluation (CPE) *task is to find the probability $P_\mathcal{B}(\bar{x} \in \rho(\mathcal{R}))$, respectively $P_\mathcal{B}(\bar{x} \in m(\varphi))$ where $m(\varphi)$ are the models (solutions) of $\varphi$).*

Alternatively, we can envision situations when one wants to assess the belief of a proposition given partial, disjunctive information.

*Belief assessment conditioned on a CNF evidence* is the task of assessing $P(X|\varphi)$ for every variable $X$. Since $P(X|\varphi) = \alpha P(X \wedge \varphi)$ where $\alpha$ is a normalizing constant relative to $X$, computing $P(X|\varphi)$ reduces to a CPE task for the query $((X = x) \wedge \varphi)$. More generally, $P(\varphi|\psi)$ can be derived from $P(\varphi|\psi) = \alpha_\varphi \cdot P(\varphi \wedge \psi)$ where $\alpha_\varphi$ is a normalization constant relative to all the models of $\varphi$.

## 4 MIXED GRAPHS AS I-MAPS

In this section we define the *mixed graph* of a mixed network and an accompanying separation criterion, extending d-separation. We show that a mixed graph is a minimal I-map (independency map) of a mixed network relative to an extended notion of separation, called *dm-separation*.

DEFINITION **4 (A mixed graph)** *Given a mixed network $M_{(\mathcal{B},\mathcal{R})}$, the mixed graph $G_M = (G_\mathcal{B}, G_\mathcal{R})$ is defined as follows. Its nodes are the set of variables $X$, and the arcs are the union of the directed arcs in the belief network graph $G_\mathcal{B}$ and the undirected arcs in the constraint graph $G_\mathcal{R}$. The moral mixed graph is the union of the moral graph of the belief network, and the constraint graph.*

The notion of *d-separation* in belief networks is known to capture conditional independence [Pearl1988]. Namely any *d-separation* in the directed graph corresponds to a conditional independence in the corresponding probability distribution. Likewise, an undirected graph representation of probabilistic networks (e.g., Markov networks) allows reading valid conditional independence based on undirected graph separation.

In this section we define *dm-separation* for mixed graphs and show that it provides a criterion for establishing minimal I-mapness for mixed networks.

DEFINITION **5 (ancestral graphs in mixed networks)** *Given a mixed graph $G_M = (G_\mathcal{B}, G_\mathcal{R})$ of a mixed network $M_{(\mathcal{B},\mathcal{R})}$ where $G_\mathcal{B}$ is the directed graph of $\mathcal{B}$, and $G_\mathcal{R}$ is the undirected constraint graph of $\mathcal{R}$, the ancestral graph of $Y \subseteq X$ in $G_M$ is the union of $G_\mathcal{R}$ and the ancestral graph of $Y$ in $G_\mathcal{B}$.*

DEFINITION **6 (dm-separation)** *Given a mixed graph, $G_M$ and given three subsets of variables $W$, $Y$ and $Z$ which are disjoint, we say that $W$ and $Y$ are dm-separated given $Z$ in the mixed graph $G_M$, denoted $\langle W, Z, Y \rangle_{dm}$, iff in the ancestral mixed graph of $W \cup Y \cup Z$, all the paths between $W$ and $Y$ are intercepted by variables in $Z$.*

THEOREM **1 (I-map)** *Given a mixed network $M = M_{(\mathcal{B},\mathcal{R})}$ and its mixed graph $G_M$, then $G_M$ is a minimal I-map relative to dm-separation. Namely, if $\langle W, Z, Y \rangle_{dm}$ then $P_M(W|Y, Z) = P_M(W|Z)$ and no arc can be removed while maintaining this property.*

**Example 1** *Figure 1(a) shows a regular belief network in which $W$ and $Y$ are d-separated given the empty set. If we add a constraint $R_{PQ}$ between $P$ and $Q$, we obtain the mixed network in Figure 1(b). According to dm-separation $W$ is no longer independent of $Y$ given the empty set, because of the path $WPQY$ in the ancestral graph. Figure 1(c) shows the auxiliary network, with variable $A$ assigned*

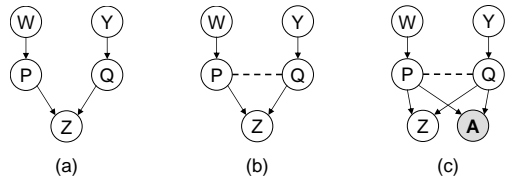

Figure 1: *dm-separation* in mixed networks

*to 1 corresponding to the constraint between P and Q. D-separation also dictates a dependency between W and Y, given A = 1.*

We will next see the first virtue of *mixed* vs *auxiliary* networks. It is now clear that the concept of constraint propagation has a well defined meaning within the mixed network framework. That is, we can allow the constraint network to be processed by any constraint propagation algorithm to yield another, equivalent, mixed network.

DEFINITION **7 (equivalent mixed networks)** *Two mixed networks defined on the same set of variables $X = \{X_1, ..., X_n\}$ and the same domains, $D_1, ..., D_n$, denoted $M_1 = M_{(\mathcal{B}_1, \mathcal{R}_1)}$ and $M_2 = M_{(\mathcal{B}_2, \mathcal{R}_2)}$, are equivalent iff they are equivalent as probability distributions, namely iff $P_{M_1} = P_{M_2}$.*

**Proposition 2** *If $\mathcal{R}_1$ and $\mathcal{R}_2$ are equivalent constraint networks (have the same set of solutions), then $M_{(\mathcal{B}, \mathcal{R}_1)}$ is equivalent to $M_{(\mathcal{B}, \mathcal{R}_2)}$.*

The above proposition shows one advantage of looking at mixed networks rather than at auxiliary networks. Due to the explicit representation of deterministic relationships, notions such as inference and constraint propagation are naturally defined and exploitable in mixed network.

## 5  AND/OR SEARCH SPACES FOR GRAPHICAL MODELS

One way of taking advantage of the implications of Proposition 2 is by search. The intuitive idea for mixed networks is to search in the space of partial variable assignments, and use the constraints to limit the actual searched space.

This sections introduces the basics of a new AND/OR search space paradigm for graphical models. The usual way to do search (called here *OR search*) is to instantiate variables in turn (in a static or dynamic ordering). In the most simple case this defines a search tree, whose nodes represent states in the space of partial assignments, and the typical depth first (DFS) algorithm searching this space would require linear space. If more space is available, then some of the traversed nodes can be cached, and retrieved when encountered again, and the DFS algorithm would in this case traverse a graph rather than a tree.

In contrast to inference algorithms which exploit the independencies in the underlying graphical model effectively (e.g. variable elimination, tree-clustering), the *OR* search space does not capture any of the structural properties of the underlying graphical model. Introducing $AND$ nodes into the OR search space can capture the graph-model structure by decomposition the problem into independent subproblems.

The *AND/OR search space* is a well known problem solving approach developed in the area of heuristic search, that exploits the problem structure to decompose the search space. The states of an AND/OR space are of two types: *OR* states which usually represent alternative ways of solving the problem, and *AND* states which usually represent problem decomposition into subproblems, all of which need to be solved. We will next present the AND/OR search space for a general *reasoning graphical model* which in particular applies to mixed networks. For more details see [Dechter2004].

For illustration consider the simple tree graphical model in Figure 2a, over domains $\{1, 2, 3\}$ which represents a graph-coloring problem. Once variable $X$ is assigned the value 1, the search space it roots corresponds to two independent subproblems, one that is rooted by $Y$ and the other rooted by $Z$. These two search subspaces do not interact. This can be captured by viewing the assignment $\langle X, 1 \rangle$ as an AND state, having two descendants. One is labeled by variable $Y$ and the other by variable $Z$. The same decomposition can be associated with the other assignments to $X$. Applying the decomposition recursively to $Y$ and $Z$ and so on along the tree (Figure 2a) yields the AND/OR search tree in Figure 2c. Notice that in the AND/OR space a full assignment to all the variables is not a path in the search space but a subtree. A solution subtree is highlighted in 2c. Clearly, the size of the AND/OR search space can be far smaller than that of the regular OR space (compare the number of states in 2b with that in 2c).

### 5.1  AND/OR SEARCH TREES

The definition of an AND/OR space is not restricted to tree graph-models, however it has to be guided by a tree which spans the original graph-model. We can use a DFS spanning tree. Given a DFS traversal of a graph $G$, the corresponding *DFS spanning tree* $T$ is defined by taking only the traversed arcs of $G$.

Given a reasoning graphical model $\mathcal{R}$, its primal graph $G$ and a DFS tree $T$ of $G$, the associated AND/OR tree is defined as follows. The *AND/OR search tree* has alternating levels of AND and OR nodes. The OR nodes are labeled $X_i$ and correspond to the variables. The AND nodes are labeled $\langle X_i, v \rangle$ and correspond to the values $v$ assigned to $X_i$. The structure of the AND/OR search tree is based on the underlying DFS tree $T$. The root of the AND/OR search tree is an OR node, labeled with the root of $T$. The children of an OR node $X_i$ are AND nodes labeled with its possible assignments $\langle X_i, v \rangle$. The children of an AND node $\langle X_i, v \rangle$ are OR nodes labeled with the children of variable $X_i$ in the DFS tree $T$. The value of leaf nodes is "S" (solved) if they represent a partial consistent assignment, or "U" if they corresponds to a dead-end.

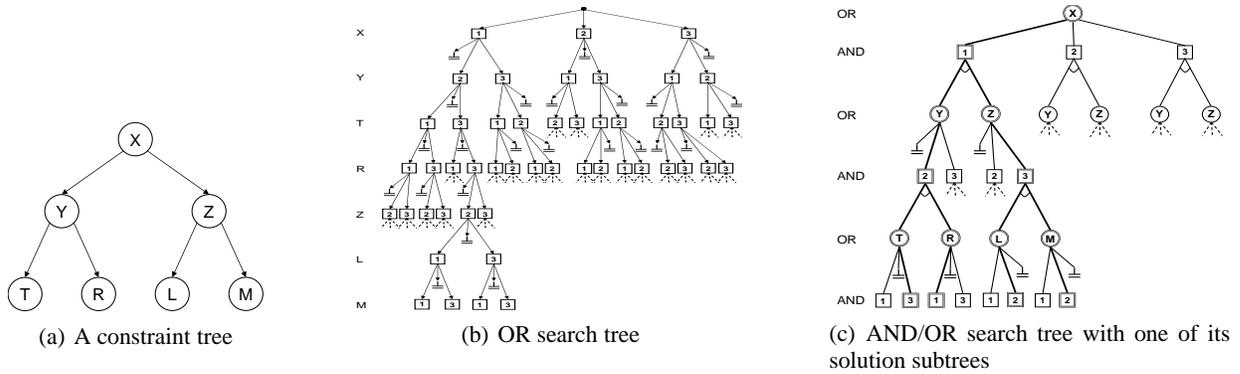

Figure 2: OR vs. AND/OR search trees; note the connector for AND arcs

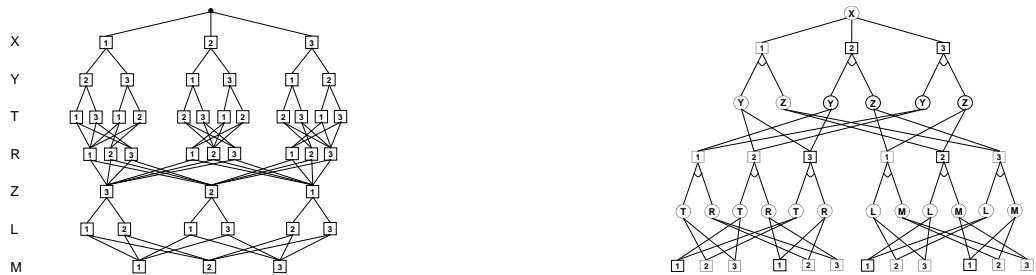

Figure 3: Condensed OR graph for the tree problem

Figure 4: AND/OR search graph for the tree problem

A **solution subgraph** of an AND/OR search graph $G$ is a subtree which: (1) contains the start node $s_0$; (2) if $n$ in the subtree is an OR node then it contains one of its child nodes in $G$ and if $n$ is an AND node it contains all its children in $G$; (3) all its terminal nodes are "Solved" (S). If we look at a probabilistic network that expresses a positive probability distribution each full assignment will be expressed as "Solved" in the AND/OR search tree.

When a depth-first search algorithm is applied to the AND/OR search tree, it requires linear space, storing only the current path from root. It is therefore important that during the search, the scope of every function from $F$ be fully assigned on some path. The DFS tree $T$ of $G$ has the property that if we add to $T$ all the other arcs of $G$ which do not appear in $T$, only back-arcs (i.e. arcs between a node and one of its ancestor) will be created. In other words no arcs will be added between different branches of $T$, which ensures that each scope of $F$ will be fully assigned on some path in $T$.

The size of the AND/OR search tree will depend on the depth of the underlying DFS tree $T$. Therefore, DFS trees of smaller height are better. However, there is a larger class of spanning trees that can be used to derive AND/OR search trees, called *legal trees*, which have the above mentioned back-arc property.

DEFINITION **8 (legal tree of a graph)** *Given an undirected graph $G = (V, E)$, a directed rooted tree $T = (V, E')$ defined on all its nodes is* legal *if any arc of $G$ which is not included in $E'$ is a back-arc, namely it connects a node to an ancestor in $T$. The arcs in $E'$ may not all be included in $E$. Given a legal tree $T$ of $G$, the* extended graph *of $G$ relative to $T$ is defined as $G^T = (V, E \cup E')$.*

Clearly, any DFS tree and any chain are legal trees. Searching the OR space corresponds to searching a chain-based space, which is a special legal tree. It is easy to see that the size of the AND/OR tree is exponential in the depth of the legal tree. Therefore, any algorithm searching this space is bounded by that complexity. Finding a legal or a DFS tree of minimal depth is known to be NP-complete. However the problem was studied, and various greedy heuristics are available. The following relationship between the induced-width and the depth of legal trees is well known [Bayardo & Miranker1996, Dechter2003]. Given a tree-decomposition of a primal graph $G$ having $n$ nodes, whose tree-width is $w^*$, there exists a legal tree $T$ of $G$ whose depth, $m$, satisfies: $m \leq w^* \cdot \log n$. In summary,

**THEOREM 2 ([Dechter2004])** *Given a graphical model $\mathcal{R}$ and a legal tree $T$, its AND/OR search tree $S_T(\mathcal{R})$ is sound and complete (contains all and only solutions) and its size is $O(n \cdot \exp(m))$ where $m$ is the legal tree's depth. A graphical model that has a tree-width $w^*$ has an AND/OR search tree whose size is $O(\exp(w^* \cdot \log n))$.*

## 5.2 AND/OR SEARCH GRAPHS

It is often the case that certain states in the search tree can be merged because the subtree they root are identical. Any two such nodes are called *unifiable*, and when merged, transform the search tree into a search graph.

### 5.2.1 Minimal AND/OR Search Graphs

A partial path in the AND/OR search-tree $S_T$ $(\langle X_1, a_1 \rangle, \langle X_2, a_2 \rangle, ..., \langle X_i, a_i \rangle)$ is abbreviated to $(\bar{X}, \bar{a}_i)$, where $\bar{X}$ is the sequence of variables and $\bar{a}$ is their corresponding sequence of value assignments.

**DEFINITION 9 (legal transformation)** *Given two partial paths over the same set of variables, $s_1 = (\bar{X}_i, \bar{a}_i)$, $s_2 = (\bar{X}_i, \bar{b}_i)$ where $a_i = b_i = v$, we say that $s_1$ and $s_2$ are unifiable at $\langle X_i, v \rangle$ (can be merged) iff the search subgraphs rooted at $s_1$ and $s_2$ are identical. The* Merge *operator over search graphs, $Merge(s_1, s_2)$ transforms $S_T$ into a graph $S'_T$ by merging $s_1$ with $s_2$.*

It can be shown that the closure under the merge operator of an AND/OR search space yields a unique fixed point,

**DEFINITION 10 (minimal AND/OR search graph)** *The minimal AND/OR search graph relative to $T$ is the closure under* merge *of the AND/OR search tree $S_T$.*

The above definition is applicable, via the legal-chain definition, to the traditional OR search tree as well, however, its compression is inferior, because of the linear structure imposed by the OR search tree. This distinction will be clarified shortly.

**Example 2** *The smallest OR search graph of the search tree in Figure 2(b) is given in Figure 3. The smallest AND/OR graph of the same problem along some DFS tree is given in Figure 4. We see that some variable-value pairs must be repeated in Figure 3 while in an AND/OR case they appear just once. For example, the subgraph below the paths $\langle X, 1 \rangle, \langle Y, 2 \rangle$ and $\langle X, 3 \rangle, \langle Y, 2 \rangle$ in Figure 3 cannot be merged.*

### 5.2.2 Rules for Merging Nodes

Given a reasoning graphical model $\mathcal{R} = (X, D, F)$ and a legal tree $T$, there could be many AND/OR graphs relative to $T$ that are equivalent to the AND/OR search tree $S_T$, each obtained by some sequence of merging. The following rules provide an efficient way for generating such graphs without creating the whole search tree first. The rules are based on a definition of *induced-width of a legal tree of $G$* which is instrumental for characterizing OR graphs vs. AND/OR *graphs*. We denote by $d_{dfs}(T)$ a DFS ordering of a tree $T$.

**DEFINITION 11 (generalized induced-width of a legal tree)** *Given $G^T$, an extended graph of $G$ relative to $T$ (see definition 9), the generalized induced width of $G$ relative to legal tree $T$, $w_T(G)$ is the induced-width of $G^T$ along $d_{dfs}(T)$.*

We can show that, 1. The *minimal* generalized induced-width of $G$ relative to all legal trees is identical to the induced-width (tree-width) of $G$. 2. The generalized induced-width of a legal chain $d$ is identical to its path-width $pw(d)$ along $d$.

Given an induced graph of $G^T$, denoted $G^{*T}$ along $d_{dfs}(T)$, each variable and its parent set is a clique.

**DEFINITION 12 (parents, parent-separators)** *Given the induced-graph, $G^{*T}$, the parents of $X$ denoted $ps_X$, are its earlier neighbors in the induced-graph. Its parent-separators, $psa_X$ are its parents that are also neighbors of future variables relative to $X$, in $T$.*

Note that for every node except those latest in the cliques of the induced graph, the parent-separators are identical to the parents. For nodes latest in cliques, the parent-separators are the separators between cliques. In $G^{*T}$, for every node $X_i$, the parent-separators of $X_i$ separates in $T$ its ancestors on the path from the root, and all its descendents in $G^T$. The reader should compare Figures 3 and 4 to verify merging using context.

**THEOREM 3** *[Dechter2004] Given $G^{*T}$, let $s_1 = (\bar{a}_i, \langle X_{i+1}, v \rangle)$ and $s_2 = (\bar{b}_i, \langle X_{i+1}, v \rangle)$ be two partial paths of assignments in its AND/OR search tree $S_T$, ending with the same assignment variable $\langle X_{i+1}, v \rangle$. If projecting $s_1$ and $s_2$ on the parent separators $psa_{i+1}$ is identical, namely: $s_1[psa_{i+1}] = s_2[psa_{i+1}]$, then the AND/OR search subtrees rooted at $s_1$ and $s_2$ are identical and $s_1$ and $s_2$ can be merged at $\langle X_{i+1}, v \rangle$.*

**DEFINITION 13 (context)** *For every state $s_i$, $s_i[psa_i]$ is called the context of $s_i$ when $psa_i$ is the parent-separators set of $X_i$ relative to the legal tree $T$.*

**THEOREM 4** *[Dechter2004] Given $G$, a legal tree $T$ and its induced width $w = w_T(G)$, the size of the AND/OR search graph based on $T$ obtained when every two nodes in $S_T$ having the same context are merged is $O(n \cdot k^w)$, when $k$ bounds the domain size.*

```
AND-OR-CPE()
Input: A mixed network M_(B,R) = (X, D, G, P, C). A
DFS tree T rooted at X_1 of the moral mixed graph of M_(B,R).
Output: The probability P(x̄ ∈ ρ(R)) that a tuple satisfies
the constraint query.
(1) Initialize OPEN by adding X_1 to it (X_1 is an OR node);
    PATH := φ
(2) if OPEN == φ
       return g(X_1)
    Remove the first node in OPEN, call it n
    Add n to PATH
(3) Expand n generating all its successors as follows:
    if (n is an OR node, denote n by X_i)
       g(X_i) := 0
       succ(X_i) := {⟨X_i, v⟩ | relevant constraints C_j, s.t.
          scope(C_j) ⊆ PATH ∪ {⟨X_i, v⟩}, are satisfied }
    else (n is an AND node, denote n by ⟨X_i, v⟩)
       g(⟨X_i, v⟩) := 1
       A := {P(Y|pa_Y) | (X_i ∈ pa_Y ∪ {Y}) and (pa_Y ∪
          {Y} ⊆ PATH)} (CPTs with fully assigned scope
          containing X_i)
       if A ≠ φ
          g(⟨X_i, v⟩) := g(⟨X_i, v⟩) * ∏_A P(Y = y | pa_Y),
       if g(⟨X_i, v⟩) == 0
          succ(⟨X_i, v⟩) := φ
       else
          succ(⟨X_i, v⟩) := Children(X_i) in T
    Add succ(n) on top of OPEN
(4) while succ(n) == φ
    (a) if (n is an OR node)
        g(Parent(n)) := g(Parent(n)) * g(n)
        if (g(n) == 0)
           remove succ(Parent(n)) from OPEN
           succ(Parent(n)) := φ
    (b) if (n is an AND node)
        g(Parent(n)) := g(Parent(n)) + g(n)
    succ(Parent(n)) := succ(Parent(n)) − {n}
    remove n from PATH
    n := Last(PATH)
(5) go to step (2)
```

Figure 5: Algorithm AND-OR-CPE

Thus, the minimal AND/OR search graph of $G$ relative to $T$ is $O(n \cdot k^w)$ where $w = w_T(G)$. Since $\min_T\{w_T(G)\}$ equals $w^*$ and since $\min_{T \in chains}\{w_T(G)\}$ equals $pw^*$,

**Corollary 1** *The minimal AND/OR search graph is bounded exponentially by the primal graph's tree-width while the OR minimal search graph is bounded exponentially by its path-width.*

It is known [Bodlaender & Gilbert1991] that for any graph $w^* \leq pw^* \leq w^* \cdot \log n$. It is also easy to place $m^*$ (the minimal depth legal tree) yielding $w^* \leq pw^* \leq m^* \leq w^* \cdot \log n$.

The difference between tree-width and path-width can be substantial. In fact for balanced trees the tree-width is 1 while the path-width is $\log n$, where $n$ is the number of variables, yielding a substantial difference between OR and AND/OR search graphs.

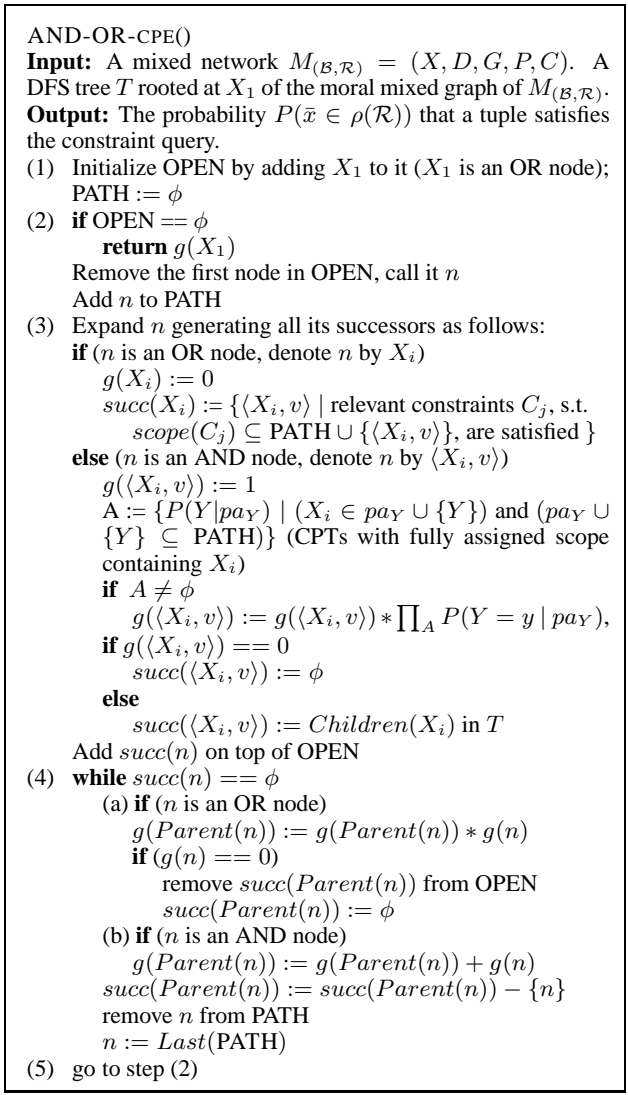

Figure 6: a) Mixed network; b) DFS tree; c)AND/OR search tree

# 6 ALGORITHMS FOR PROCESSING MIXED NETWORKS

We will focus on the CPE task of computing $P(\bar{x} \in \rho(\mathcal{R}))$, the probability that a random tuple satisfies the constraint query. A number of related tasks can be easily derived by changing the appropriate operator (e.g., using maximization for maximum probable explanation - MPE, or summation and maximization for maximum a posteriori hypothesis - MAP).

There are two primary exact approaches for processing belief and constraint networks: inference and search. Both of them can be applied in the context of the mixed networks. Variable elimination algorithms were explored in [Dechter & Larkin2001]. The experimental work of [Dechter & Larkin2001] demonstrated that keeping the deterministic information separately was far superior to embedding it in the auxiliary network.

Variable elimination algorithms are expected to be far better than linear space search, as is predicted by worst-case complexity. Yet, for large or highly connected networks, variable elimination may be infeasible due to space limitations. Algorithms with controllable space are the only ones applicable in such situations. They use less space at the cost of spending more time.

## 6.1 LINEAR SPACE ALGORITHM OF AND/OR SEARCH TREES

We will present first the extreme case, a new linear space algorithm based on depth first search for processing mixed networks. The algorithm explores the AND/OR search trees just introduced.

The algorithm, AND-OR-CPE, is described in Figure 5. It is given as input a legal tree $T$ of the mixed moral graph, and the output is the result of the CPE task, the probability that a random tuple satisfies the constraint query. AND-OR-CPE traverses the AND/OR search tree corresponding to $T$ in a DFS manner. Each node maintains a label $g$ which accumulates the computation resulted from its subtree. OR nodes accumulate the summation of their chil-dren's labels, while AND nodes accumulate the product of their children's labels.

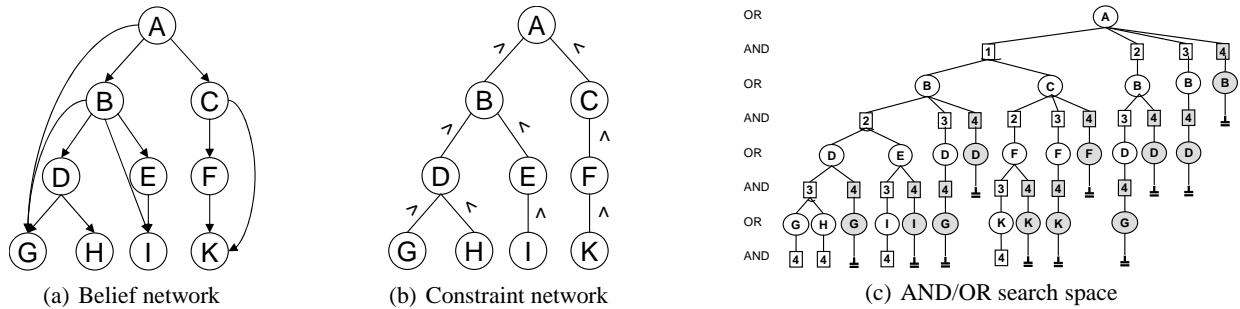

Figure 7: Example of AND-OR-CPE and AO-FC search spaces

A list called OPEN simulates the recursion stack. The list PATH maintains the current assignment. $Parent(n)$ refers to the predecessor of $n$ in PATH, which is also its parent in the AND/OR tree, and $succ$ denotes the set of succesors of a node in the AND/OR tree.

Step (3) is where the search goes forward. When an OR node is expanded, it is labeled with 0, and its successors are the values that are consistent with the current assignment. To determine these successors, only the relevant constraints, whose scope is contained in the current path, need to be checked. When an AND node $\langle X_i, v \rangle$ is expanded, it is labeled with the product of all the CPT entries for which $X_i$ is contained in their scope, and the scope is contained in PATH (i.e., it is fully assigned). If the product does not exist, the label is 1.

Step (4) is where the labels are propagated backward. This is triggered when a node has an empty set of successors, and it typically happens when the node's descendants are all evaluated or when it is a dead-end.

**Example 3** *Figure 6(a) shows a mixed binary network (the constraint part is given by the cnf formula $\varphi$). Figure 6(c) describes an AND/OR search tree based on the DFS tree given in Figure 6(b). Algorithm* AND-OR-CPE *starts from node A, and assigns $g(A) = 0$, then $g(\langle A, 0 \rangle) = P(A{=}0)$. It continues assigning $g(C) = 0$, and then $g(\langle C, 0 \rangle) = 1$. B is not assigned yet, so $P(C|A, B)$ will participate in the label of a descendant node (the set A of step (3) of the algorithm is empty). The node D can take both values ($\varphi$ is not violated), so by backing up the values of its descendents $g(D)$ becomes 1 ($g(D) = \sum_D P(D|C{=}0) = 1$). Going on the branch of B, $g(B) = 0$, then B can only be extended to 0 (to satisfy $A \vee \neg B$), and the label becomes $g(\langle B, 0 \rangle) = P(B{=}0) \cdot P(C{=}0|A{=}0, B{=}0)$. In general, a CPT participates in labeling at the highest level (closer to the root) of the tree where all the variables in its scope are assigned.*

The following are implied immediately from the general properties of AND/OR search trees,

**THEOREM 5** *Algorithm* AND-OR-CPE *is sound and exact for the CPE task.*

**THEOREM 6** *Given a mixed network $M$ with $n$ variables with domain sizes bounded by $k$ and a legal tree $T$ of depth $m$ of its moral mixed graph, the time complexity of* AND-OR-CPE *is $O(n \cdot k^m)$.*

**Proposition 3** *A mixed network having induced width $w^*$ has an AND/OR search tree whose size is $O(\exp(w^* \cdot \log n))$.*

#### 6.1.1 Constraint propagation in AND-OR-CPE

Proposition 2 provides an important justification for using mixed networks as opposed to auxiliary networks. The constraint portion can be processed by a wide range of constraint processing techniques, both statically before AND/OR search or dynamically during AND/OR search. The algorithms can combine consistency enforcing (arc-, path-, i-consistency) before or during search, directional consistency, look-ahead techniques, no-good learning etc.

In the empirical evaluation, we used two forms of constraint propagation on top of AND-OR-CPE (called AO-C for shortness). The first, yielding algorithm AO-FC, is based on *forward checking*, which is one of the weakest forms of propagation. It propagates the effect of a value selection to each future uninstantiated variable separately, and checks consistency against the constraints whose scope would become fully instantiated by just one such future variable. To perform this, we need to add at step (3) of Figure 5:

　　Apply forward-checking for $PATH \cup \langle X_i, v \rangle$
　　If inconsistent then do not include $\langle X_i, v \rangle$ in $succ(X_i)$

The second algorithm we used is called AO-RFC, and performs a variant of *relational forward checking*. Rather than checking only constraints whose scope becomes fully assigned, AO-RFC checks all the existing constraints by looking at their projection on the current path. If the projection is empty an inconsistency is detected. AO-RFC is computationally more intensive than AO-FC, but its search space is smaller.

Table 1: AND/OR space vs. OR space

| N=25, K=2, R=2, P=2, C=10, S=3, t=70%, 20 instances, w*=9, h=14 | | | | |
|---|---|---|---|---|
| | Time | Nodes | Dead-ends | Full space |
| AO-C | 0.15 | 44,895 | 9,095 | 152,858 |
| OR-C | 11.81 | 3,147,577 | 266,215 | 67,108,862 |

Table 2: AND/OR Search Algorithms (1)

| N=40, K=2, R=2, P=2, C=10, S=4, 20 instances, w*=12, h=19 | | | | | | | | | | | |
|---|---|---|---|---|---|---|---|---|---|---|---|
| t | i | Time AO- | | | Nodes (*1000) AO- | | | Dead-ends (*1000) AO- | | | #sol |
| | | C | RC | RFC | C | FC | RFC | C | FC | RFC | |
| 20 | 0 | 0.671 | 0.056 | 0.022 | 153 | 4 | 1 | 95 | 3 | 1 | 2E+05 |
| | 6 | 0.479 | 0.055 | 0.022 | 75 | 3 | 1 | 57 | 3 | 1 | |
| | 12 | 0.103 | 0.044 | **0.016** | 17 | 2 | 1 | 3 | 2 | 0 | |
| 40 | 0 | 2.877 | 0.791 | 1.094 | 775 | 168 | 158 | 240 | 40 | 36 | 8E+07 |
| | 6 | 1.409 | 0.445 | 0.544 | 183 | 35 | 32 | 107 | 28 | 24 | |
| | 12 | 0.189 | **0.142** | 0.149 | 28 | 9 | 7 | 3 | 4 | 3 | |
| 60 | 0 | 6.827 | 4.717 | 7.427 | 1,975 | 1,159 | 1,148 | 362 | 163 | 159 | 6E+09 |
| | 6 | 2.809 | 2.219 | 3.149 | 347 | 184 | 180 | 151 | 89 | 86 | |
| | 12 | **0.255** | **0.331** | 0.425 | 36 | 23 | 22 | 3 | 5 | 5 | |
| 80 | 0 | 14.181 | 14.199 | 21.791 | 4,283 | 3,704 | 3,703 | 370 | 278 | 277 | 1E+11 |
| | 6 | 5.305 | 6.286 | 9.061 | 626 | 519 | 518 | 128 | 98 | 97 | |
| | 12 | **0.318** | 0.543 | 0.714 | 44 | 40 | 40 | 1 | 3 | 3 | |
| 100 | 0 | 23.595 | 27.129 | 41.744 | 7,451 | 7,451 | 7,451 | 0 | 0 | 0 | 1E+12 |
| | 6 | 8.325 | 11.528 | 16.636 | 957 | 957 | 957 | 0 | 0 | 0 | |
| | 12 | **0.366** | 0.681 | 0.884 | 51 | 51 | 51 | 0 | 0 | 0 | |

Table 3: AND/OR Search Algorithms (2)

| t | i | Time | | Nodes (*1000) | | Dead-ends (*1000) | | #sol |
|---|---|---|---|---|---|---|---|---|
| | | AO-FC | AO-RFC | AO-FC | AO-RFC | AO-FC | AO-RFC | |
| N=100, K=2, R=10, P=2, C=30, S=3, 20 instances, w*=28, h=38 | | | | | | | | |
| 10 | 0 | **1.743** | **1.743** | 15 | 15 | 15 | 15 | 0 |
| | 10 | 1.748 | 1.746 | 15 | 15 | 15 | 15 | |
| | 20 | 1.773 | 1.784 | 15 | 15 | 15 | 15 | |
| 20 | 0 | **3.193** | 3.201 | 28 | 28 | 28 | 28 | 0 |
| | 10 | 3.195 | 3.200 | 28 | 28 | 28 | 28 | |
| | 20 | 3.276 | 3.273 | 28 | 28 | 28 | 28 | |
| 30 | 0 | 69.585 | 62.911 | 805 | 659 | 805 | 659 | 0 |
| | 10 | 69.803 | **62.908** | 805 | 659 | 805 | 659 | |
| | 20 | 69.275 | 63.055 | 805 | 659 | 687 | 659 | |
| N=100, K=2, R=5, P=3, C=40, S=3, 20 instances, w*=41, h=51 | | | | | | | | |
| 10 | 0 | 1.251 | 0.382 | 7 | 2 | 7 | 2 | 0 |
| | 10 | 1.249 | **0.379** | 7 | 2 | 7 | 2 | |
| | 20 | 1.265 | 0.386 | 7 | 2 | 7 | 2 | |
| 20 | 0 | 22.992 | **15.955** | 164 | 113 | 163 | 111 | 0 |
| | 10 | 22.994 | 15.978 | 162 | 110 | 162 | 111 | |
| | 20 | 22.999 | 16.047 | 162 | 110 | 162 | 110 | |
| 30 | 0 | 253.289 | 43.255 | 2093 | 351 | 2046 | 304 | 0 |
| | 10 | 254.250 | **42.858** | 2026 | 283 | 2032 | 289 | |
| | 20 | 253.439 | 43.228 | 2020 | 278 | 2026 | 283 | |

Table 4: AND/OR Search vs. Bucket Elimination

| t | i | Time | | | Nodes (*1000) | | Dead-ends (*1000) | | #sol |
|---|---|---|---|---|---|---|---|---|---|
| | | BE | AO-FC | AO-RFC | AO-FC | AO-RFC | AO-FC | AO-RFC | |
| N=70, K=2, R=5, P=2, C=30, S=3, 20 instances, w*=22, h=30 | | | | | | | | | |
| 40 | 0 | 26.4 | 2.0 | 1.3 | 49 | 21 | 35 | 19 | 0 |
| | 10 | | 1.9 | **1.2** | 30 | 18 | 29 | 18 | |
| | 20 | | 1.9 | 1.3 | 26 | 17 | 21 | 16 | |
| 50 | 0 | | 30.7 | 35.6 | 2,883 | 2,708 | 1,096 | 1,032 | 1E+12 |
| | 10 | | 18.6 | 18.9 | 557 | 512 | 342 | 302 | |
| | 20 | | 12.4 | **12.1** | 245 | 216 | 146 | 130 | |
| 60 | 0 | | 396.8 | 511.4 | 51,223 | 50,089 | 13,200 | 12,845 | 7E+14 |
| | 10 | | 167.9 | 182.5 | 5,881 | 5,708 | 2,319 | 2,241 | |
| | 20 | | **80.5** | 83.6 | 1,723 | 1,655 | 718 | 697 | |
| N=60, K=2, R=5, P=2, C=40, S=3, 20 instances, w*=23, h=31 | | | | | | | | | |
| 40 | 0 | 67.3 | 0.7 | **0.6** | 9 | 9 | 8 | 7 | 0 |
| | 10 | | 0.6 | 0.6 | 6 | 5 | 5 | 5 | |
| | 20 | | 0.6 | 0.6 | 5 | 5 | 4 | 4 | |
| 50 | 0 | | 3.2 | 3.0 | 58 | 55 | 41 | 38 | 6E+04 |
| | 10 | | 3.0 | 2.8 | 31 | 28 | 28 | 25 | |
| | 20 | | 2.7 | **2.6** | 25 | 23 | 20 | 18 | |
| 60 | 0 | | 65.2 | 70.2 | 2,302 | 2,292 | 1,206 | 1,195 | 8E+08 |
| | 10 | | 54.1 | 56.4 | 791 | 781 | 660 | 649 | |
| | 20 | | **39.6** | 40.7 | 459 | 449 | 319 | 309 | |

**Example 4** *Figure 7 shows the search spaces of* AO-C *and* AO-FC. *Figure 7(a) shows the belief part of the mixed network, and Figure 7(b) the constraint part. All variables have the same domain, {1,2,3,4}, and the constraints express "less than" relations. Figure 7(c) shows the search space of* AO-C *(the whole tree) and* AO-FC *(the grey nodes are pruned in this case).*

## 7   EMPIRICAL EVALUATION

We ran our algorithms on mixed networks generated randomly uniformly given a number of input parameters: $N$ - number of variables; $K$ - number of values per variable; $R$ - number of root nodes for the belief network; $P$ - number of parents for a CPT; $C$ - number of constraints; $S$ - the scope size of the constraints; $t$ - the tightness (percentage of the allowed tuples per constraint). (N,K,R,P) defines the belief network and (N,K,C,S,t) defines the constraint network. We report the time in seconds, number of nodes expanded and number of dead-ends encountered (in thousands), and the number of consistent tuples of the mixed network ($\#sol$). In tables, $w^*$ is the induced width and $h$ is the height of the legal tree.

We compared four algorithms: 1) AND-OR-CPE, denoted here AO-C; 2) AO-FC and 3) AO-RFC (described in previous section); 4) BE - bucket elimination (which is equivalent to join tree clustering) on the auxiliary network; the version we used is the basic one for belief networks, without any constraint propagation and any constraint testing. For the search algorithms we tried different levels of caching, denoted in the tables by $i$ (i-bound, this is the maximum scope size of the tables that are stored). $i = 0$ stands for linear space search. Caching is implemented based on context as described in Section 5.

Table 1 gives a brief account for our choice of using AND/OR space instead of the traditional OR space. Given the same ordering, an algorithm that only checks constraints (without constraint propagation) always expands less nodes in the AND/OR space.

Tables 2, 3, and 4 show a comparison of the linear space and caching algorithms exploring the AND/OR space. We ran a large number of cases and this is a typical sample.

Table 2 shows a medium sized mixed network, across the full range of tightness for the constraint network. For linear space ($i = 0$), we see that more constraint propagation helps for tighter networks ($t = 20$), AO-RFC being faster than AO-FC. As the constraint network becomes loose, the effort of AO-RFC does not pay off anymore. When almost all tuples become consistent, any form of constraint propagation is not cost effective, AO-C being the best choice in such cases ($t = 80, 100$). For each type of algorithm, caching improves the performance. We can see the general trend given by the bolded figures.

Table 3 shows results for large mixed networks ($w^* = 28, 41$). These problems have an inconsistent constraint portion ($t = 10, 20, 30$). AO-C was much slower in this case, so we only include results for AO-FC and AO-RFC. For the smaller network ($w^* = 28$), AO-RFC is only slightly better than AO-FC. For the larger one ($w^* = 41$), we see that more propagation helps. Caching doesn't improve either of the algorithms here. This means that for these inconsistent problems, constraint propagation is able to detect many of the no-goods easily, so the overhead of caching cancels out its benefits (only no-goods can be cached for inconsistent problems). Note that these problems are infeasible for BE due to high induced width.

Table 4 shows a comparison between search algorithms and BE. All instances for $t < 40$ were inconsistent and the AO algorithms were much faster than BE, even with linear space. Between $t = 40 - 60$ we see that BE becomes more efficient than AO, and may be comparable only if AO is given the same amount of space as BE.

There is an expected trend with respect to the size of the traversed space and the dead-ends encountered. We see that the more advanced the constraint propagation technique, the less nodes the algorithm expands, and the less dead-ends it encounters. More caching also has a similar effect.

## 8   CONCLUSION

The paper presents the new framework of *mixed networks* which combines belief and constraint networks. It allows for a more efficient and flexible exploitation of probabilistic and deterministic information by borrowing the specific strengths of each formalism that it builds upon. This separation is harder to exploit when constraints are expressed as CPTs. We also introduce the AND/OR search space for graphical models, which is always more effective than the traditional OR space [Dechter2004]. We demonstrate the benefit of searching the AND/OR space for solving mixed networks, by introducing a new linear space search algorithm. The AND/OR algorithm can easily be augmented with caching, to take advantage of the amount of space available.

An alternative main approach based on variable elimination was explored earlier in [Dechter & Larkin2001]. Related work was presented recently in [Allen & Darwiche2003], where unit resolution can speed up recursive conditioning [Darwiche1999] in the case of genetic linkage networks which contain a lot of determinism. In general, the recursive conditioning type algorithms exhibit behavior and have complexities similar to AND/OR search algorithms.

Overall we showed that belief networks algorithms can benefit from the mixed representation in a number of ways: 1) Constraint propagation techniques can be applied straightforwardly, maintaining their properties of convergence and fixed point; 2) The semantics is much clearer by separating probabilistic and deterministic information; 3) The algorithms can be made more efficient.

**Acknowledgments**

This work was supported in part by the NSF grant IIS-0086529 and the MURI ONR award N00014-00-1-0617.